# Tracking Algorithm for Microscopic Flow Data Collection

Kardi Teknomo[1], Yasushi Takeyama[2], Hajime Inamura[3]

**Introduction**

Various methods to automate traffic data collection have recently been developed by many researchers. A macroscopic data collection through image processing has been proposed [1]. For microscopic traffic flow data, such as individual speed and time or distance headway, tracking of individual movement is needed. The tracking algorithms for pedestrian or vehicle have been developed to trace the movement of one or two pedestrians based on sign pattern [2], and feature detection [3,4]. No research has been done to track many pedestrians or vehicles at once. This paper describes a new and fast algorithm to track the movement of many individual vehicles or pedestrians.

**Tracking Algorithm**

We have slices of images and each slice may contain several points that represent objects (vehicles or pedestrian). Each object is represented by one point in position $(x, y)$ at slice $s$. Moreover, each point represents the centroid of an area of an object in an image slice. The object can enter the screen, change position, and go out of the screen. Starting from the first slice, each object is numbered by a unique object-number. The objective of the tracking algorithm is to match the points between slices by giving an object number to each point in each slice. Two points are matched if and only if the two points represent one object. Since each point in each slice denote a location at a time, the distance between two match points in two consecutive slices can also represent the speed of that object. Matrix $D^s$ is a binary matrix that represents the distance between point $i$ in slice $s$ and point $j$ in slice $s+1$ and

$$D^s(i, j) = \begin{cases} 1 \leftrightarrow d_{ij} \leq T \\ 0 \leftrightarrow d_{ij} > T \end{cases}$$

Where,
T is the threshold between objects in a slice. $d_{ij}$ = Distance between points $i$ in slice $s$ and $j$ in slice. If q is number of points in slice $s$ and $r$ is number of points in slice $s+1$, then $D^s$ is matrix $q$ by $r$. If the distance between points in two slices is the smallest and it is smaller than the threshold, then the two points are the same object, or

$$\min_{\forall j} (d_{ij} \leq T) \Leftrightarrow \{i, j\} \in G_i.$$

In other words, we assumed that distance between objects is bigger than the distance between points in two consecutive slices. This assumption is acceptable if the number of frame/second is quite large. Assuming a linear relationship between speed and density, then the number of frames per second depend on the free flow speed, $\mu_f$, and the gradient between the speed and density of an object, b, is

$$n > \frac{\mu_f^2}{4b}$$

Binary distance matrix $D^s$ has the following properties:
- If all the entries in a row $i$ are zeros, then that the object $i$ in slice $s$ has no match in slice $s+1$. Object $i$ will *go out* of the screen at slice $s+1$.
- If all the entries in a column $j$ are zeros, then that the object $j$ in slice $s+1$ has no match in slice $s$. Object $j$ is a *newcomer* to the screen at slice $s+1$.
- If one point represents one object, and there is no occlusion, then the row element and column element of matrix $D^s$ has only one entry, which is 1.

Based on those properties, the matching algorithm can be described mathematically as follow:
1. For slice $s = 1$: $G_i^s = i$ and $H = \{G_i^s\}$
2. For slice $s = 2$ to max Slice:
   a. $\sum_i D^{s-1}(i, j) = 0 \Rightarrow G_j^s = \{\max(H) + 1\}, H = \{H \cup G_j^s\}$

---

[1] Doctoral Student, Graduate School of Information Sciences, Tohoku University Japan
[2] Associate Professor, Graduate School of Information Science, Tohoku University Japan
[3] Professor, Graduate School of Information Science, Tohoku University Japan

b. $\sum_j D^{s-1}(i, j) = 0 \Rightarrow H = \{H - G_i^{s-1}\}$

c. $\sum_i D^{s-1}(i, j) = 1$ and $\sum_j D^{s-1}(i, j) = 1 \Rightarrow \{G_j^s = G_i^{s-1} | G_j^s \in H\}$

Where $H$ as a set of object-numbers that is still on the screen. $G_i^s$ is a set of object-numbers at slice $s$ row $i$. $G_j^s$ is set of object-numbers at slice $s$ column $j$. For the first time, the object-number is an ordered number, similar to the row of matrix $D^1$. For the second slice up to the end of the stack, the algorithm is as follow:

- If all entries of the column vector of matrix $D^{s-1}$ are zeros, put an object-number in slice $s$, $G_i^s$, as the maximum number in set $H$ plus one, and then put this new $G_i^s$ into set $H$.
- If all entries of the row vector of matrix $D^{s-1}$ are zeros, remove the object-number in slice s-1 row $i$, $G_i^{s-1}$, from set $H$.
- If there is only one entry in matrix $D^{s-1}$ row $i$ column $j$, put the object-number of row $j$ in slice $s$, $G_j^s$, as the object-number of row i in slice $s-1$, $G_i^{s-1}$, where the $G_j^s$ is also subset of $H$.

**Experiments Result and Discussion**

The above algorithm was implemented on real world car and pedestrian traffic. Video taking was recorded with a fixed focus, at the top of the road or walkway to avoid projection and occlusion between objects. The video was converted to files using a freeware "NIH image" [5,6] at a rate about 30 frames/second.

Background image is a slice where the road or walkway does not have any vehicles or pedestrians. Image difference between the background image and each slice was performed. A macro program was developed to remove the background, to filter the images, to threshold and analyze particles to get the centroid of each object. After that, the tracking algorithm as described above was executed.

The result of the tracking algorithm is matrix of object numbers, and its coordinates at each time slice. Calibration into real world coordinates and traffic flow variables calculation are then performed. These calculations are beyond the scope of this paper. By comparing the automatic and manual data collection, it reveals that the algorithm is working without error, as long as there is no occlusion between vehicles or pedestrians. Occlusion is a phenomenon wherein two or more objects may become too close to each other and form a point and then later these points may separate again as individual points. The condition without occlusion can be obtained as long as the traffic density is light to medium. Since the shadow of objects may make the object occlude choice of location and light source also needs to be considered.

**Conclusion**

Tracking algorithm to trace the coordinates of individual movements of vehicles or pedestrians on the road or walkway was proposed. It was found that moving vehicles or pedestrians are successfully tracked for light to medium traffic density. Further research to overcome the occlusion problems is recommended.